%% file: SIURO_minor_revised_arXiv.tex
\documentclass{article}
\usepackage[utf8]{inputenc}
\usepackage{graphics}
\usepackage{tikz}

\usepackage[square,sort,comma,numbers]{natbib}

\usepackage{placeins}
\usepackage[top=0.5in, bottom=0.75in, left=1in, right=1in]{geometry}
\usepackage{multirow}
\usepackage{titlesec}
\usepackage{soul}
\usepackage{mathtools}
\usepackage{subcaption}
\usepackage[labelformat=parens,labelsep=quad,skip=3pt]{caption}

\setcitestyle{square}
\usepackage[mathscr]{euscript}

\input{math_commands.tex}

\newcommand{\cL}{{\mathcal{L}}}

\newcommand{\myfootnote}[1]{
\renewcommand{\thefootnote}{}
\footnotetext{\hspace{-16.5pt}\footnotesize#1}
\renewcommand{\thefootnote}{\arabic{footnote}}}

\include{header}

\begin{document}
\title{Analysis of Legal Documents via Non-negative Matrix Factorization Methods}

\author{Ryan Budahazy\textsuperscript{\rm 1}, 
        Lu Cheng\textsuperscript{\rm 2}, 
        Yihuan Huang\textsuperscript{\rm 2}, 
        Andrew Johnson\textsuperscript{\rm 3},
        Pengyu Li\textsuperscript{\rm 2}, \\
        Joshua Vendrow\textsuperscript{\rm 2}, 
        Zhoutong Wu\textsuperscript{\rm 2},
        Denali Molitor\textsuperscript{\rm 2},
        Elizaveta Rebrova\textsuperscript{\rm 2}, 
        Deanna Needell\textsuperscript{\rm 2}\\
}

\myfootnote{\noindent     \textsuperscript{\rm 1} Department of Mathematics, Towson University, Towson, MD \\
    \textsuperscript{\rm 2} Department of Mathematics, University of California, Los Angeles, CA \\
    \textsuperscript{\rm 3} Department of Mathematics, Baruch College, New York, NY \\
    Email addresses: ryanbudahazy@gmail.com (Ryan Budahazy), lucheng@g.ucla.edu (Lu Cheng), charlotte0408@g.ucla.edu (Yihuan Huang), ajohnson0764@gmail.com (Andrew Johnson), erby1215@g.ucla.edu (Pengyu Li), jvendrow@math.ucla.edu (Joshua Vendrow), zhoutongwu@ucla.edu (Zhoutong Wu), deanna@math.ucla.edu (Prof. Deanna Needell).
}

\definecolor{joshcolor}{rgb}{0.2, 0.6, 0.4}
\newcommand{\commjv}[1]{{\textcolor{joshcolor}{(#1 --JV)}}} 

\newcommand{\jcobl}[1]{\textcolor{red}{[#1 -- Jacob]}}

\definecolor{dncolor}{rgb}{0.4, 0.6, 0.5}
\newcommand{\dn}[1]{\textcolor{dncolor}{[#1 -- dn]}}

\maketitle

\begin{abstract}
    
The California Innocence Project (CIP), a clinical law school program aiming to free wrongfully convicted prisoners, evaluates thousands of mails containing new requests for assistance and corresponding case files. Processing and interpreting this large amount of information presents a significant challenge for CIP officials, which can be successfully aided by topic modeling techniques. In this paper, we apply Non-negative Matrix Factorization (NMF) method and implement various offshoots of it to the important and previously unstudied data set compiled by CIP. We identify underlying topics of existing case files and classify request files by crime type and case status (decision type). The results uncover the semantic structure of current case files and can provide CIP officials with a general understanding of newly received case files before further examinations. We also provide an exposition of popular variants of NMF with their experimental results and discuss the benefits and drawbacks of each variant through the real-world application.  
\end{abstract}

\section{Introduction}


The California Innocence Project (CIP), founded in 1999, is a clinical law school program hosted by California Western School of Law. Its core mission is to free wrongfully convicted prisoners, with the additional goals of reforming the criminal justice system and training upcoming law students. Since its inception, it has freed over 30 people who were wrongly incarcerated in the past years.

After many years of operation, CIP has collected a significant quantity of unprocessed data. CIP receives approximately 1,500 new requests for assistance and over 4,000 pieces of digitized mail annually. This amount of information presents a challenge for the volunteers at CIP who manually process or classify these case files. At the same time, it also offers an opportunity for data analysis that can reveal important information for CIP to understand the cases themselves and help CIP to evaluate its own decision process in handling cases.

In this paper, we utilize Non-negative Matrix Factorization (NMF) method \citep{lee1999learning} and various offshoots of it to cluster CIP case files data and generate meaningful topics. Given that the nature of CIP data is mostly document based, after converting those documents (PDFs of handwriting) into texts, we can identify significant words that appear in those texts to form coherent lexical topics. Analyzing topics generated by each method allows us to answer various questions and address challenges that CIP has such as categorizing old case files, automating the categorization of incoming requests for assistance, and perhaps most importantly understanding why certain cases were chosen to pursue as opposed to others.


Classical NMF \citep{lee1999learning} helps us understand large amount of data by offering a low-rank representation of a data set. Semantic NMF \citep{ailem2017non} can take advantage of the hidden contextual and semantic meanings from a data set by training word-context dictionaries. Hierarchical NMF (HNMF) \citep{kuang2013fast,tu2018hierarchical,gao2019neural} can uncover the hierarchical structure of a data set at varying granularities by providing factorizations of the data set at many different ranks and identifying the relationship between the factorizations. Supervised NMF (SNMF) and semi-supervised NMF (SSNMF) \citep{hyekyoung_lee_semi-supervised_2010} add a regularization term that utilizes class label information, the crime and decision type of each case, to form class-distinct topics and can also be used for classification.

\subsection{Contributions}

In this paper, we utilize a variety of Non-negative Matrix Factorization methods in order to identify salient information within case files from the California Innocence Project. The main contributions of this paper are as follows: 


    
    
    
    


\begin{itemize}

    \item We investigate an important and previously unstudied data set of case files from the California Innocence Project, and provide a pre-processing pipeline that can be applied to similar types of data.
    
    \item We identify meaningful lexical topics within the CIP data set using various Non-negative Matrix Factorization methods that help us to uncover hidden semantic structure, and to predict the crime type and case status of unlabeled and new case files. 
    
    \item We provide an exposition of popular variants of NMF, and analyze the experimental difference observed from applying these variants to a real-world data set.  
\end{itemize}

\subsection{Organization}

In Section \ref{sec:data}, we describe the data pre-processing steps and the construction of the CIP data matrix. In Section \ref{sec:methods},  relevant works and algorithms of various NMF methods are introduced for performing topic modeling  and classification. Section \ref{sec:experiments} contains corresponding experimental results from the application and implementation of those methods, such as discovery of latent topics or classification accuracy. In the following Section \ref{sec:Conclusion},  we compare and contrast different methods through the perspective of how each of their results can assist CIP officials. In the same section, we discuss potential improvements and suggest some future directions. 


\FloatBarrier
\section{Data} \label{sec:data}

\subsection{Description of the Data set}

The raw data set consists of over 1,000 case files provided by CIP. All case files are stored as PDF documents. Each case contains documents such as letters, applications, questionnaires, and other legal instruments from a single inmate. In the paper, we analyze the content of 3 different documents for each case:
\begin{enumerate}
    \item \textit{Initial Letter}\label{Word: IL}: The first document CIP receives from an inmate is an initial letter requesting assistance. Some of them are short and direct while others offer details of the case and the reason for their innocence. Analysis of these letters is important because they are the first document CIP screens.
    \item \textit{Appellant's Opening Brief (AOB)} \label{Word: AOBs}: A document written by the appellant to argue for their innocence and explain the mistakes made by the court in their decision. This document contains important information about the facts and potential evidence of the case. 
    \item \textit{Questionnaire} \label{Word: QN}: To collect basic information, all inmates requesting help from CIP are required to fill out a standardized questionnaire. We use this questionnaire to identify labels such as convicted crime type. 
\end{enumerate}

\noindent After examining the documents in each case file, CIP officials split all case files into six decision types: 

\begin{enumerate}
    \item \textit{Letters Requesting Assistance}: This type contains cases with only an initial letter.
    \item \textit{Cases for Consideration}: This type contains cases with sufficient information to begin a process of deciding whether to pursue the case.
    \item \textit{Cases for Investigation}: This type contains cases that have been through CIP's vetting process and are ready to be
assigned to a clinic student for additional investigation.
    \item \textit{Cases to be Closed}: This type contains cases that CIP has determined do not warrant investigation.
    \item \textit{Unresponsive Files}: This type contains cases where the inmate become unresponsive.
    \item \textit{Cases we have won}: This type contains cases that have been pursued and won by CIP, meaning that the inmate's innocence has been proved before the court.
\end{enumerate}

\noindent For our methods, we utilize information from two decision types: the \textit{Cases for Investigation} and \textit{Cases to be Closed}. In total, \textit{Cases for Investigation} has $55$ cases with initial letters and $63$ with AOBs. \textit{Cases to be Closed} has $169$ cases with initial letters and $214$ with AOBs. There are $42$ cases in \textit{Cases for Investigation} and $93$ cases in \textit{Cases to be Closed} with both initial letters and AOBs.

 


\subsection{Data Pre-Processing} \label{sec:preprocessing}

Here we detail the pre-processing steps taken to convert the raw CIP data into a data matrix  formatted as input into the methods described in Section \ref{sec:methods}.

\bigbreak
\noindent\textbf{Optical Character Recognition}.
All of our models require raw text as input, so we perform Optical Character Recognition (OCR) to extract text from the PDF documents, using Google Cloud's Vision API. This task proves especially difficult for \textit{Initial Letters} and the \textit{Questionnaires}, which are all handwritten.

\bigbreak
\noindent \textbf{Spellchecking}.
Once documents are converted to raw text, we perform spellchecking using the \texttt{pyspellchecker} PyPi package. 
These errors are especially prevalent in the handwritten responses, where spelling errors can come from both mistakes by the writers of the text and the OCR process. We also perform basic cleaning such as removing all non-word characters. 

\bigbreak
\noindent\textbf{Tf-idf}.
Following the works of \citep{ramos2003using} and \citep{li2007keyword}, we apply term-frequency inverse document frequency (tf-idf) \citep{SALTON1988513} to represent each document by a vectorized bag-of-words representation. Tf-idf is a numerical statistic that reflects the importance of a word to a document in a collection. For this bag-of-words representation, we also remove basic stopwords according to the NLTK English stopwords list \citep{nltk}, and remove names of inmates. Note that different methods use different tuning parameters to build tf-idf vocabulary and the parameters will be described in Section \ref{sec:experiments}. 

\begin{figure}[H]
    \centering
    \includegraphics[width=0.6\textwidth]{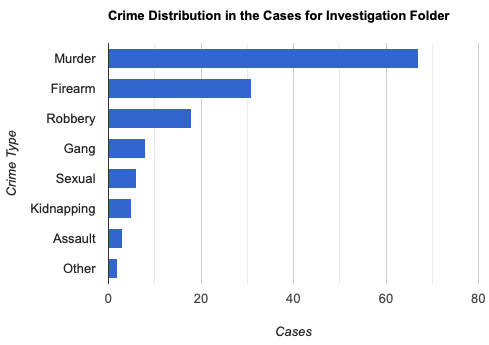}
  \caption{The distribution of convicted crime for cases within \textit{Cases for Investigation} type.}
  \label{fig:distribution}
\end{figure}

\bigbreak
\noindent\textbf{Data Labeling}.\label{subsec:labels}
In preparation for applying supervised models, we first extract the decision type for each case file, \textit{Cases for Investigation} or \textit{Cases to be Closed}.
Secondly, we extract crime type from the corresponding \textit{questionnaires} for each case file as its label information. We examine answers to the question in the questionnaire corresponding to each case, ``List all of the crimes for which you are currently serving time, and where the conviction was entered", and extract answers as crime type labels for each case correspondingly. Each case can be associated with one or more of the following crime types: \textit{assault}, \textit{drug}, \textit{gang}, \textit{kidnapping}, \textit{murder}, \textit{robbery}, \textit{sexual}, \textit{vandalism}, \textit{manslaughter}, \textit{theft}, \textit{burglary} and \textit{stalking}. Figure \ref{fig:distribution} shows the distribution for crime type in the \textit{Cases for Investigation} type. We can see that the crime types in our data set are mainly violent crimes, such as murder.  




\section{Methods and Related Works} \label{sec:methods}

Non-negative Matrix Factorization (NMF) method was first introduced in \citep{lee1999learning} for semantic analysis of text documents and documents clustering; see also \citep{gillis_why_2014, 4053063, xu2003document}. Since then, many variants of NMF methods have been proposed. In this section, we review the functionality and algorithm details for the NMF variants we applied to our data set. After the data pre-processing, we construct a data matrix in the form of $X = [x_1, x_2, \cdots, x_n] \in \R_{\geq 0}^{d \times n}$, where $d$ is the number of words in our vocabulary generated by tf-idf and $n$ is the number of documents. Each entry $X_{ij}$ is the tf-idf of the $i^{th}$ word in the $j^{th}$ document. We use the standard notation for the Frobenious matrix norm, $\|A\|_F^2 = \sum_{i,j}A^2_{i,j}$.

\subsection{Non-negative Matrix Factorization (NMF)} \label{subsec:NMF_method}
Given a rank $r$, which also can be interpreted as the number of desired topics, the classical NMF method approximates and decomposes the matrix $X$ into 2 low-rank non-negative matrices $W = [w_1, w_2, \cdots, w_r] \in \R_{\geq 0}^{d \times r}$, the dictionary matrix, and $H = [h_1, h_2, \cdots, h_3] \in \R_{\geq 0}^{r \times n}$, the coding matrix \citep{lee1999learning}. The matrices $W$ and $H$ can be found by solving the optimization problem: 


\begin{equation}\label{eq: basic_NMF_eq}
\operatorname*{argmin}_{W\in \mathbb{R}^{d\times r}_{\ge 0},\, H\in \mathbb{R}^{r\times n}_{\ge 0} }  \lVert X - WH  \rVert_{F}^{2}.
\end{equation}
Each column of $W$, $w_i \in \R^{d}$, contains the weight for each word in the $i^{th}$ topic and each column of $H$, $h_j \in \R^{r}$, contains the weight for each topic in the $j^{th}$ document. By analyzing each $w_i$ and $h_j$, we are able to represent each topic by its top weighted words as well as assign each document to its most related topic, namely the one with the largest corresponding weight. 


\subsection{Semantic NMF} \label{subsec:Semantic_NMF_method}
Classical realizations of NMF use ``bag-of-words" document representations, such as the tf-idf representation (Section \ref{sec:preprocessing}), that do not account for the sequential order in which words occur in documents. This could result in a significant loss of semantic contexts. Semantic NMF \citep{ailem2017non} was developed to overcome this shortcoming of using a ``bag-of-words" representation in NMF. 
In order to avoid the loss, a word-context matrix $C \in \mathbb{R}_{+}^{d \times d}$ is computed such that each row $i$ represents a word $w_i$ in our vocabulary and each column $j$ is the corresponding context word $w_j$. The value for each entry $c_{i,j}$ is obtained by counting the number of times the word pair ($w_i$, $w_j$) appeared in a $l$-length window among all the documents \citep{ailem2017non}. Semantic NMF leverages the semantic relationships between words and takes  context into consideration by introducing a new matrix $M \in \R ^ {d \times d}$ which is known as the Shifted Positive Point-wise Mutual Information (SPPMI) matrix \citep{Levy2014NeuralWE}. Each entry of the SPPMI matrix measures the connection between word $w_i$ and its context word $w_j$ with each entry given by:
\begin{equation}
m_{i, j} = \max\{{\log\left(\frac{c_{i,j} \times c_{.,.}}{c_{i,.} \times c_{.,j}}\right)-\log(N),0}\},
\end{equation}
where $c_{\cdot, \cdot} = \sum_{j, j^{'}}{c_{j, j^{'}}}$, $c_{j, \cdot} = \sum_{j^{'}}{c_{j, j^{'}}}$, and $c_{\cdot,  j^{'}} = \sum_{j}{c_{j, j^{'}}}$. $N \in \R$ is the constant controlling the shift, and
on a small data set choosing $N$ to be $5$ is proven to perform well \citep{kenyondean2019word}. The matrices $W$ and $H$ can be found by solving the following optimization problem: 
\begin{equation} \label{eq:SemanticNMF}
    \operatorname*{argmin}_{W\in \mathbb{R}^{d\times r}_{\ge 0},\, H\in \mathbb{R}^{r\times n}_{\ge 0} }\underbrace{\frac{1}{2} \norm{X-WH}_F^2}_\text{classical NMF}  
    + \underbrace{\frac{1}{2} \norm{M-WSW^T}_F^2}_\text{word embedding},
\end{equation} 
where the matrix $S \in \R ^{r \times r}_{\ge 0}$ is a square symmetric matrix that offers additional degrees of freedom \citep{4053063} and plays a key role when $M$ is approximated and decomposed. The first term denoted as classical NMF captures 
how words form a document while the latter term denoted as word embedding captures the co-occurrence information between words \citep{ailem2017non}. The resulting $W$ and $H$ matrices can be interpreted in the same way for classical NMF, as described in the end of Section \ref{subsec:NMF_method}.

\subsubsection{Semantic NMF with Keyword Highlighting} \label{subsubsec:Semantic_NMF_KH_method}

In order to encourage certain ``important'' words within our vocabulary to play a greater role in the factorization, we propose a method we refer to as \textit{keyword highlighting} for encouraging \textit{a priori} keywords to take part in the topic modeling. This simple method allows us to supervise the formation of topics relevant to the CIP, and uncover more hidden latent information in the data set that does not appear following a standard NMF. To perform keyword highlighting, we choose a subset of words from the vocabulary, which correspond to rows in the data matrix $X$, and directly multiply these rows by a large constant factor. We find that when combining keyword highlighting with semantic NMF, we are able to form precise and meaningful topics that match the desired a priori keywords we wish to participate in the topic modeling.

\subsection{Hierarchical NMF (HNMF)} 
Hierarchical NMF is an extension of NMF that seeks to elucidate hierarchical structure in a data set. Two classical versions of HNMF differ in how they reveal the hierarchical structure, top-down \citep{kuang2013fast} and bottom-up \citep{gao2019neural}. 
Top-down hierarchical NMF allows us to recursively split topics of a low granularity into topics of a larger granularity.
On the other hand, bottom-up hierarchical NMF helps us identify topics of a lower granularity by combining together topics from the factorization of a larger granularity.

\subsubsection{Top-down HNMF Algorithm} \label{subsubsec:TD_HNMF_method}
The top-down HNMF algorithm begins by performing classical NMF on the data matrix X using rank $r^{(1)}$
\begin{equation}
    X \approx W^{(0)}H^{(0)},
\end{equation}
to obtain $r^{(1)}$ super-topics in the first layer with the dictionary matrix $W^{(0)}$ and the coding matrix $H^{(0)}$ of the first layer. According to the coding matrix $H^{(0)}$, columns of $X$ are then split into sub-matrices $X_1^{(1)}, X_2^{(1)}, \cdots, X_{r^{(1)}}^{(1)}$, each representing documents within a topic. Then, classical NMF is applied to each new matrix $X_1^{(1)}, X_2^{(1)}, \cdots, X_{r^{(1)}}^{(1)}$
\begin{align}
    \begin{split}
    X_{1}^{(1)} &\approx W_{1}^{(1)}H_{1}^{(1)}, \\ 
    X_{2}^{(1)} &\approx W_{2}^{(1)}H_{2}^{(1)}, \\
    & \vdots \\
    X_{r^{(1)}}^{(1)} &\approx W_{r^{(1)}}^{(1)}H_{r^{(1)}}^{(1)},
    \end{split}
\end{align}
to obtain corresponding dictionary and coding matrices of the second layer. 
The $(i + 1)^{th}$ layer can be built by repeating this process for $X_{r^{(i)}}^{(i)}$, where $r^{(i)}$ is the rank of matrix $W_{r^{(i)}}^{(i)}$.  

\subsubsection{Bottom-up HNMF Algorithm} \label{subsubsec:BU_HNMF_method}
Given a series of decreasing ranks $k^{(0)}, k^{(1)}, \ldots, k^{(\cL-1)}$, the bottom-up HNMF provides a factorization at each of the $\cL$ ranks, and by relating the topics provided at each rank we can understand how sub-topics combine into super-topics. To perform bottom-up HNMF for $\cL$ layers, we factor the data matrix as
\begin{align}
    \begin{split}
    X &\approx W^{(0)}H^{(0)}, \\ 
    X &\approx W^{(0)}W^{(1)}H^{(1)}, \\
    & \vdots \\
    X &\approx W^{(0)}W^{(1)}\cdots W^{(\cL)}H^{(\cL)}.
    \end{split}
\end{align}
where $W^{(i)}$ represents the how the sub-topics at layer $i$ collect into the super-topics at layer $i+1$. 

\FloatBarrier

\subsection{Supervised NMF and Semi-supervised NMF}
The semi-supervised NMF method (SSNMF) and supervised NMF (SNMF) incorporate information from known class labels into NMF in order to predict unknown labels, taking advantage of the topic information discovered by NMF \citep{hyekyoung_lee_semi-supervised_2010}. 
Given the data matrix $X$, the associated class label matrix is defined as $Y=[y_1, \cdots, y_n]\in \mathbb{R}_{\geq 0}^{p\times n}$, where $p$ is number of classes and $n$ is number of documents. Each column $y_i$ is an encoding vector such that the $j^{th}$ entry of $y_i$ is $1$ if the document $i$ belongs to label class $j$, and otherwise is $0$.

\subsubsection{Supervised NMF} \label{subsubsec: SNMF_method}
To perform supervised NMF, the data matrix $X$ is partitioned into training data $X_\mathrm{train} \in \mathbb{R}^{d\times m}$ and testing data $X_\mathrm{test} \in \mathbb{R}^{d\times (n - m)}$ with corresponding label matrices $Y_\mathrm{train}\in \mathbb{R}^{p\times m}$ and $Y_\mathrm{test}\in \mathbb{R}^{p\times (n - m)}$, where $m$ is the number of documents in the training data.
The dictionary matrix $W_\mathrm{train}$, coding matrix $H_\mathrm{train}$, and label dictionary matrix $B_\mathrm{train}$ can be found by solving the optimization problem:

\begin{equation} \label{eq:SNMF}
\operatorname*{argmin}_{W_\mathrm{train}\in \mathbb{R}^{d\times r}_{\ge 0},\, H_\mathrm{train}\in \mathbb{R}^{r\times m}_{\ge 0}, \, B_\mathrm{train}\in \mathbb{R}^{p\times r}_{\ge 0}}\|X_\mathrm{train}-W_\mathrm{train}H_\mathrm{train}\|^2_F+\lambda\|(Y_\mathrm{train}-B_\mathrm{train}H_\mathrm{train})\|_F^2,
\end{equation}
where $\lambda$ is a regularization parameter determining the importance of the supervised term. The resulting $W_\mathrm{train}$ and $H_\mathrm{train}$ matrices can be interpreted in the same way for classical NMF, as described in the end of Section \ref{subsec:NMF_method}
and matrix $B_\mathrm{train}$ can be interpreted as the dictionary matrix for the label matrix $Y_\mathrm{train}$.  
\bigbreak

\subsubsection{Semi-supervised NMF} \label{subsubsec: SSNMF_method}

Instead of partitioning the data matrix $X$ into training data $X_\mathrm{train}$ and testing data $X_\mathrm{test}$,
semi-supervised NMF utilizes a masking matrix $L=[l_1,\dots,l_n]\in\mathbb{R}_+^{p\times n}$ to set some documents' label information as unknown and exclude the information from training the model. The masking matrix $L$ is defined as
\begin{equation}
    l_j = \begin{cases}
    \textbf{1}_k, & \text{if the label of $x_j$ is known}\\
    \textbf{0}_k, & \text{otherwise}
    \end{cases} 
\end{equation}
where $\textbf{1}_k = [1,\dots,1]^T\in \mathbb{R}^p$ and $\textbf{0}_k = [0,\dots,0]^T\in \mathbb{R}^p$.

The dictionary matrix $W$, coding matrix $H$, and label dictionary matrix $B$ can be found by solving the following optimization problem: 
\begin{equation}\label{eq:10}
\operatorname*{argmin}_{W\in \mathbb{R}^{d\times r}_{\ge 0},\, H\in \mathbb{R}^{r\times n}_{\ge 0}, \, B\in \mathbb{R}^{p\times r}_{\ge 0}} \|X-WH\|^2_F+\lambda\|L\odot(Y-BH)\|_F^2, 
\end{equation}
where $A \odot B$ denotes the entry-wise multiplication between matrix $A$ and $B$. The resulting $W$, $H$, and $B$ matrices can be interpreted in the same way for supervised NMF. It is important to note that when semi-supervised NMF is used to predict the labels for the testing matrix, the information from the $WH$ decomposition for the testing matrix is available, while it is not when supervised NMF is used to make such predictions. During the training, semi-supervised NMF takes $X_\mathrm{test}$ into consideration, while supervised NMF only focuses on $X_\mathrm{train}$.    


\subsubsection{Labeling Accuracy Score} \label{subsubsec:LAS}
In general, both semi-supervised NMF and supervised NMF utilize a portion of the label matrix $Y$ to reconstruct/predict the label information for the other portion. In semi-supervised NMF, the reconstructed portion of label matrix $Y$, namely $Y^\prime$, can be found by: 

\begin{equation}\label{eq:reco_ssNMF_y}
 Y^\prime = (J_{p, n} - L)\odot(BH)
\end{equation}
where $J_{p, n} \in \R^{p \times n}$ is an all-ones matrix. In supervised NMF, the reconstructed portion $Y_\mathrm{test}$ can be found by: 

\begin{equation}
    Y_\mathrm{test}=B_\mathrm{train}(W_\mathrm{train}^TW_\mathrm{train})^{-1}W_\mathrm{train}^T X_\mathrm{test}.
\end{equation}

By comparing the label information in the reconstructed portion, $Y^\prime$ and $Y_\mathrm{test}$, with the actual label of that portion, we design the following metrics to measure how successfully both models can predict/assign label information. We first set the largest element in each column of $Y^\prime$ and $Y_\mathrm{test}$ to be $1$ and all other entries to be $0$. This means that we only regard the most related class label as the label being predicted. A document's label prediction is regarded as successful if the predicted label matches one of the actual labels for that document. The Labeling Accuracy Score (LAS) is then defined and calculated as the percentage of total number of successful predictions divided by the total number of documents in testing data. 



\FloatBarrier

\section{Results} \label{sec:experiments}



In this section, we discuss the topics discovered by applying NMF from the \texttt{sklearn} package \citep{sklearn_api} and implementing semantic NMF, semantic NMF with keyword highlighting, and HNMF on the initial letters and AOBs. We also include visualizations of reconstructed crime type and decision type label matrices by SSNMF and SNMF as well as the corresponding Labeling Accuracy Scores. 


\FloatBarrier
\subsection{Discovering Topics in Initial Letters} \label{subsec:Result_IL}

We first focus on discovering topics within the Initial Letter data set that combines all initial letters from both the \textit{Cases for Investigation} and \textit{Cases to be Closed} decision types. 
The number of topics (the rank for the factorization) is chosen experimentally based on which number yields the most coherent and consistent topics. When analyzing the Initial Letter data set, we empirically found that using a rank 7 factorization provides us with clear and diverse topics for all methods.  We also build different vocabularies for different methods to optimize the topic results. The tf-idf vocabulary for classical NMF is created using tuning parameters ``max\_df=$0.8$" and ``min\_df=$0.015$" in the function ``TfidfVectorizer"; the tf-idf vocabulary for semantic NMF is created using tuning parameters ``max\_{df}=$0.8$", ``min\_{df}=$0.01$", and ``max\_features=$500$"; the tf-idf vocabulary for hierarachical NMF is created using tuning parameters ``max\_{df}=$0.8$", ``min\_{df}=$0.04$".


\FloatBarrier
\subsubsection{Classical NMF} \label{subsubsec: NMF_IL}
In Table \ref{tab:NMF_InitialLetterTopics}, we display the topic keywords formed by classical NMF (Section \ref{subsec:NMF_method}) on the initial letters. Topic $1$ is related to courtroom trial proceeding for each case. Analyzing documents in this topic can potentially identify cases whose initial letters contain more comprehensive basic information. Other interesting topics may be Topic $6$ and Topic $7$, which include a collection of different types of evidence mentioned in the initial letters. Topic $6$ describes police investigation that involves an eye witness and Topic $7$ contains information about physical evidence such as DNA or blood. We notice that all Spanish words get attributed to Topic 3 and this suggests that there are documents in Spanish in our data set and that they were not clustered in any meaningful way. 
This is a shortcoming of our implementation that a collection of multilingual corpora is not able to be handled properly; however, Topic $3$ does help identify the bilingual feature of the CIP data set and automatically separates the Spanish sub-corpus for future analysis.  We will discuss in detail how to handle those Spanish case files in Section \ref{sec:Conclusion}.


\begin{table}[htbp]
\centering
\begin{tabular}{lllllll}
\hline
\textbf{Topic 1} & \textbf{Topic 2} & \textbf{Topic 3} & \textbf{Topic 4} & \textbf{Topic 5} & \textbf{Topic 6} & \textbf{Topic 7} \\ \hline
trial     & help      & que     & said  & would         & witness     & dna       \\
evidence  & need      & por     & told  & like          & police      & blood     \\
attorney  & please    & gracias & got   & thank         & suspect     & apartment \\
jury      & know      & yo      & get   & send          & trial       & fact      \\
defendant & innocence & eston   & would & innocence     & description & victim    \\
testimony & crime     & swedes  & went  & questionnaire & interview   & items     \\
judge     & years     & su      & going & screening     & also        & profile   \\
never     & convicted & es      & car   & concern       & said        & done      \\
sentence  & prove     & para    & never & dear          & gave        & could     \\
years     & hello     & mucha   & asked & address       & detective   & detective \\ \hline
\end{tabular}
    \caption{The top $10$ keywords learned by NMF on the initial letters from both categories}
	\label{tab:NMF_InitialLetterTopics}%
\end{table}

\FloatBarrier
\subsubsection{Semantic NMF} \label{subsubsec: Semantic_NMF_IL}
In Table \ref{tab:semantic_letters}, we display the topic keywords discovered by implementing semantic NMF (Section \ref{subsec:Semantic_NMF_method}) on the Initial Letter data set. 
Those topics are hard to interpret due to the lack of a single focus in one topic or missing relevant information for understanding. For example, both Topic $5$ and Topic $7$  describe shooting incidents. Topics $1$ and $3$ are related to signing different forms, indicated by keywords ``signed" in Topic $1$ and ``petition" in Topic $3$, yet the detailed information about those forms are not revealed under these topics.
To address issues of lacking focus and low interpretability, we then highlight a set of keywords provided by CIP by increasing their weight by a factor of $1.5$ in the data matrix $X$. The majority of highlighted keywords are related to physical evidence and the full set is attached in the Appendix \ref{subsec:keywords Highlighted}.


In Table \ref{tab:semantic_key_letters}, we display the topic keywords formed by semantic NMF with keyword highlighting (Section \ref{subsubsec:Semantic_NMF_KH_method}) on the Initial Letter data set. We find that the topic results are more specific when compared with semantic NMF without keyword highlighting. 
Topic $5$ focuses on evidence such as camera footage; other keywords, such as "driving" or "car", in this topic can also be associated with traffic incidents. Topic $7$ focuses on general evidence that might appear in the courtroom during appellate such as eyewitness's claim or transcripts.

\begin{table}[H]
    \centering
    \begin{tabular}{ccccccc}
    \Xhline{2\arrayrulewidth}
		\textbf{Topic 1} & \textbf{Topic 2} & \textbf{Topic 3} & \textbf{Topic 4} & \textbf{Topic 5} & \textbf{Topic 6} & \textbf{Topic 7}\\
	\hline 
pages & words & por & get & getting & written & hands \\
writ & writing & que & motion & incident & reports & told \\
reasonable & request & sentencing & like & shot & warrant & ther \\
serving & commit & july & program & entire & detective & shot \\
respectfully & arrested & yo & appointed & done & phone & innocence \\
ineffective & wrote & direction & prove & fiction & thats & body \\
informed & contacted & es & turn & police & yes & suppose \\
january & pay & expert & sending & knowledge & notice & decided \\
signed & preliminary & petition & lived & parents & knew & december \\
non & respond & looking & send & happened & sign & brother \\
    \Xhline{2\arrayrulewidth}
    \end{tabular}
        \caption{The top $10$ keywords learned by semantic NMF on the initial letters from both categories}
    \label{tab:semantic_letters}
\end{table}

\begin{table}[htb]
    \centering
    \begin{tabular}{ccccccc}
    \Xhline{2\arrayrulewidth}
		\textbf{Topic 1} & \textbf{Topic 2} & \textbf{Topic 3} & \textbf{Topic 4} & \textbf{Topic 5} & \textbf{Topic 6} & \textbf{Topic 7}\\
	\hline 
related & gringo & que & cdc & driving & tattoo & misconduct \\
villalobos & soon & yo & ba & camera & pages & ineffective \\
camarena & july & por & october & feet & list & claim \\
garage & sincerely & es & respectfully & cars & recovered & preliminary \\
cerberus & wrong & direction & appreciate & footage & began & transcript \\
believed & fingerprint & attention & inmate & beat & using & eyewitness \\
helped & obtain & expert & bless & women & previous & appellate \\
pulled & possible & attempted & helping & minutes & photo & statements \\
step & wrote & petition & began & according & strike & actual \\
members & however & detectives & hopes & completely & recording & bring \\
    \Xhline{2\arrayrulewidth}
    \end{tabular}
        \caption{The top $10$ keywords learned by semantic NMF with keyword highlighting on the initial letters from both categories}
    \label{tab:semantic_key_letters}
\end{table}

\FloatBarrier
\subsubsection{HNMF} \label{subsubsec: HNMF_IL}
For both the top-down (Section \ref{subsubsec:TD_HNMF_method}) and bottom-up (Section \ref{subsubsec:BU_HNMF_method}) HNMFs, we begin from an initial factorization into 7 super-topics (rank 7). For top-down HNMF, each of the 7 super-topics in the first layer is split into three sub-topics. As a result, we discover a total of 21 sub-topics from the initial letters. In bottom-up HNMF, the seven-topic layer is regarded as the sub-topic layer, and 2 super-topic layers are built to combine the 7 sub-topics. 

In Table \ref{tab:HNMF_topdown_initial_letter}, we display the topic keywords in each of the two layers formed by top-down HNMF on the Initial Letter data set. 
We can see that most top keywords associated with the first-layer super-topics are not evenly distributed into three sub-topics. Rather, they are clustered in one sub-topic. For example, important court-related keywords of Super-topic 1 all appear in Sub-topic 1 (``trial", ``evidence", ``testimony"). 

In Tables \ref{tab:hnmf_letter_7} through \ref{tab:hnmf_letter_3}, we display a three layer bottom-up HNMF on the Initial Letter data set with topic numbers 7, 5 and 3. We see that the hierarchical factorization not only preserves unique topic related to trails (Topic 1 in Tables \ref{tab:hnmf_letter_7} through \ref{tab:hnmf_letter_3}) but also combines overlapping topics in the first layer related to evidence and police procedure (Topic 2 and Topic 3 in Table \ref{tab:hnmf_letter_7}), and combines topics related to seeking assistance from CIP (Topics 5 through 7). 
We also noticed that the last topic, which is a list of Spanish words, is preserved across three layers. This might suggest the existence of a considerable amount of Spanish documents in the text corpora. If we would include them in our topic modeling algorithm, a new pipeline that targets specifically Spanish texts is needed. Some alternative ways to handle multilingual data is further discussed in Section \ref{sec:Conclusion}.

\begin{table}[H]
    \centering
    \resizebox{\textwidth}{!}{
    \begin{tabular}{p{5cm}|p{4cm}p{4cm}p{4cm}}
    \Xhline{2\arrayrulewidth}
		\textbf{Primary Topics} & \textbf{Subtopic 1} & \textbf{Subtopic 2} & \textbf{Subtopic 3}\\
	\hline 
\textbf{Super-topic 1}: \\ 
trial, evidence, attorney, jury, testimony, defendant, filed, judge, sentence, never
&
\textit{trial}, \textit{evidence}, \textit{testimony}, \textit{jury}, witnesses, \textit{never}, witness, \textit{attorney}, testify, testified
&
\textit{filed}, petition, review, \textit{trial}, \textit{defendant}, charges, areas, innocence, \textit{attorney}, denied
&
years, crime, also, \textit{attorney}, murder, could, help, \textit{sentence}, would, transcripts
\\
\hline
\textbf{Super-topic 2}: \\ 
help, please, need, know, innocence, crime, years, convicted, hello, let
&
\textit{help}, \textit{know}, \textit{hello}, \textit{innocence}, \textit{crime}, \textit{convicted}, really, cip, \textit{please}, sincerely
&
year, robbery, matter, \textit{need}, prove, ano, evidence, sentence, \textit{years}, concern
&
\textit{help}, \textit{please}, \textit{need}, send, get, thank, waiting, innocent, bless, sa
\\
\hline
\textbf{Super-topic 3}: \\ 
que, por, es, direction, expert, petition, much, favor, sa, oct
&
\textit{que}, \textit{por}, \textit{direction}, \textit{es}, \textit{petition}, \textit{favor}, \textit{oct}, \textit{much}, done, \textit{sa}
&
cd, ay, \textit{que}, evidence, proof, presented, parole, video, bring, te
&
\textit{por}, \textit{que}, \textit{expert}, mailed, november, cdc, legal, want, \textit{es}, attorney
\\
\hline
\textbf{Super-topic 4}: \\ 
said, got, told, get, would, going, went, never, car, asked
&
\textit{told}, \textit{said}, attorney, \textit{asked}, trial, \textit{would}, police, \textit{car}, \textit{going}, wife
&
\textit{got}, \textit{get}, hoping, yrs, \textit{going}, people, innocent, fight, \textit{went}, change
&
guilty, \textit{would}, could, report, help, know, innocence, like, evidence, also
\\
\hline
\textbf{Super-topic 5}: \\ 
would, thank, like, innocence, send, questionnaire, concern, screening, request, dear
&
\textit{innocence}, conviction, te, please, write, evidence, sincerely, years, also, crime
&
provide, application, interested, \textit{concern}, appreciated, \textit{thank}, writing, currently, wanted, possible
&
\textit{would}, \textit{like}, \textit{send}, \textit{questionnaire}, \textit{thank}, \textit{screening}, \textit{concern}, address, \textit{innocence}, please
\\
\hline
\textbf{Super-topic 6}: \\ 
witness, police, suspect, interview, trial, description, said, also, crime, detective
&
\textit{witness}, \textit{police}, \textit{said}, \textit{interview}, also, \textit{trial}, \textit{description}, gave, murder, \textit{suspect}
&
people, \textit{crime}, que, te, ay, like, might, evidence, ther, cd
&
testified, \textit{trial}, victim, officer, officers, attorney, \textit{said}, count, told, filed
\\
\hline
\textbf{Super-topic 7}: \\ 
dna, blood, fact, done, stated, detective, could, also, items, victim
&
trial, \textit{also}, attorney, te, conviction, evidence, crime, statement, shot, count
&
need, help, like, please, send, know, dear, thank, hear, october
&
\textit{dna}, \textit{blood}, \textit{fact}, \textit{done}, \textit{detective}, \textit{victim}, said, \textit{stated}, know, \textit{could}\\
    \Xhline{2\arrayrulewidth}
    \end{tabular}
}
     \caption{Topic keywords learned by Top-down HNMF on the initial letters from both categories. Keywords in primary topics are italicized in subtopics. }
      \label{tab:HNMF_topdown_initial_letter}
\end{table}

\begin{table}[htb]
    \centering
    \begin{tabular}{ccccccc}
    \Xhline{2\arrayrulewidth}
		\textbf{Topic 1} & \textbf{Topic 2} & \textbf{Topic 3} & \textbf{Topic 4} & \textbf{Topic 5} & \textbf{Topic 6} & \textbf{Topic 7}\\
	\hline 
trial & dna & witness & said & help & would & que \\
evidence & blood & police & told & need & like & por \\
attorney & apartment & suspect & got & please & thank & gracias \\
jury & fact & trial & get & know & send & yo \\
defendant & items & description & would & innocence & innocence & eston \\
testimony & victim & interview & went & crime & questionnaire & swedes \\
judge & profile & also & going & years & screening & su \\
never & done & said & car & convicted & concern & es \\
sentence & could & gave & never & prove & dear & para \\
years & detective & detective & asked & hello & address & mucha \\
    \Xhline{2\arrayrulewidth}
    \end{tabular}
    \caption{The Top $10$ topic keywords learned by the first layer of Bottom-up HNMF on the initial letters from both categories}
    \label{tab:hnmf_letter_7}
\end{table}

\begin{figure}[h]
\begin{minipage}{0.5\textwidth}
    \centering
    \begin{tabular}{ccccc}
    \Xhline{2\arrayrulewidth}
		\textbf{Topic 1} & \textbf{Topic 2} & \textbf{Topic 3} & \textbf{Topic 4} & \textbf{Topic 5}\\
	\hline 
trial & witness & help & would & que \\
evidence & said & need & like & por \\
attorney & told & please & thank & gracias \\
jury & got & know & send & yo \\
defendant & police & innocence & innocence & eston \\
never & also & years & questionnaire & swedes \\
testimony & get & crime & screening & su \\
judge & never & convicted & concern & es \\
sentence & dna & prove & dear & para \\
years & would & let & address & mucha \\
    \Xhline{2\arrayrulewidth}
    \end{tabular}
        \captionof{table}{The Top $10$ topic keywords learned by the second layer of Bottom-up HNMF on the initial letters from both categories}
    \label{tab:hnmf_letter_5}
\end{minipage}
\hfill
\begin{minipage}{0.3\textwidth}
    \centering
    \begin{tabular}{ccc}
    \Xhline{2\arrayrulewidth}
		\textbf{Topic 1} & \textbf{Topic 2} & \textbf{Topic 3}\\
	\hline 
trial & help & que \\
evidence & please & por \\
attorney & innocence & gracias \\
never & need & yo \\
witness & know & eston \\
said & thank & swedes \\
also & convicted & su \\
would & crime & es \\
jury & would & para \\
police & years & mucha \\
    \Xhline{2\arrayrulewidth}
    \end{tabular}
    \captionof{table}{The Top $10$ topic keywords learned by the third layer of bottom-up HNMF on the initial letters from both categories} 
    \label{tab:hnmf_letter_3}
\end{minipage}
\end{figure}

\FloatBarrier
\subsection{Discovering Topics in AOBs} \label{subsec: Topics_AOB}

We next focus on discovering topics within the Appellant's Opening Brief (AOB) data set which combines all AOBs from both the \textit{Cases for Investigation} and \textit{Cases to be Closed} decision types. The topic number is set to 10 for all methods in this section since 10 as the topic number consistently gives us clear and diverse topics from the AOB data set. As the AOBs are typically longer than the initial letters, presumably containing more content, it makes sense to summarize more topics from AOBs. We also build different vocabularies for different methods to optimize the topic results. The tf-idf vocabularies for classical NMF and hierarchical NMF are created using the tuning parameters ``max\_{df}=$0.8$" and ``min\_{df}=$0.04$" in the function ``TfidfVectorizer"; and the tf-idf vocabulary for semantic NMF is created using the tuning parameters ``max\_features = $700$".

\FloatBarrier
\subsubsection{Classical NMF}\label{subsubsec: NMF_AOB}
In Table \ref{tab:nmf_aob}, we display the topic keywords by classical NMF (Section \ref{subsec:NMF_method}) on the AOBs. We can see that most topics successfully capture one specific type of crime. For example, Topic $2$ can be associated with illegal gang activities. Topic $3$ can be associated with general murder cases; Topic $6$ focuses on murder cases involving multiple accomplices; while Topic $8$ is related to murder cases involving gun shooting. Topic $5$ provides contextualizing information for cases involving sexual assaults and Topic $10$ can be associated with burglary cases.

\begin{table}[htb]
    \centering
    \resizebox{\textwidth}{!}{
    \begin{tabular}{cccccccccc}
    \Xhline{2\arrayrulewidth}
		\textbf{Topic 1} & \textbf{Topic 2} & \textbf{Topic 3} & \textbf{Topic 4} & \textbf{Topic 5} & \textbf{Topic 6} & \textbf{Topic 7}& \textbf{Topic 8}& \textbf{Topic 9} & \textbf{Topic 10}\\
	\hline 
prosecutor    & gang        & murder        & strike      & sexual      & murder      & juror         & car       & suggestive      & burglary    \\
misconduct    & members     & manslaughter  & enhancement & sex         & accomplice  & jurors        & phone     & witness         & apartment   \\
witness       & member      & passion       & injury      & rape        & aider       & prospective   & detective & eyewitness      & intent      \\
prejudicial   & expert      & voluntary     & felony      & duress      & abettor     & misconduct    & shooting  & photo           & mayhem      \\
statements    & shooting    & heat          & bodily      & vagina      & robbery     & deliberations & apartment & lineup          & residential \\
prejudice     & crips       & killing       & robbery     & touching    & special     & motion        & murder    & identifications & threat      \\
admission     & gangs       & premeditation & discretion  & lewd        & intent      & verdict       & shot      & shooting        & instruction \\
prosecutorial & murder      & malice        & strikes     & penis       & abetting    & dna           & officer   & photographic    & felony      \\
delay         & enhancement & provocation   & imposed     & offenses    & aiding      & excused       & going     & suspect         & burglaries  \\
witnesses     & car         & deliberation  & firearm     & penetration & instruction & instruction   & got       & pack            & unanimity   \\ \cline{1-7}

    \Xhline{2\arrayrulewidth}
    \end{tabular}
    }
    \caption{The top 10 keywords learned by NMF on AOBs from both categories}
\label{tab:nmf_aob}
\end{table}

\FloatBarrier
\subsubsection{Semantic NMF} \label{subsubsec: Semantic_NMF_AOB}
In Table \ref{tab:semantic_aob}, we display the topic keywords by semantic NMF (Section \ref{subsec:Semantic_NMF_method}) on AOBs. Those topics are hard to interpret due to the overlapping across multiple topics and the ambiguity of topic words. For example, Topic $1$ and Topic $10$ can both be understood as court related, due to keywords ``trial" and ``evidence" in Topic $1$ as well as ``transcript" and ``respondent" in Topic $10$. And Topics $3$ and $4$ contain ambiguous keywords such as ``object", ``generally", ``come", or ``going", which provide little information for understanding. To address the overlapping  and ambiguous topic issues, we increase the weights of a set of keywords provided by CIP 
by a half of their original weights. The full set of keywords is attached in the Appendix \ref{subsec:keywords Highlighted}.

In Table \ref{tab:semantic_key_aob}, we display the topic keywords by semantic NMF with keyword highlighting (Section \ref{subsubsec:Semantic_NMF_KH_method}) on  AOBs. The result contains novel and diverse topics. Also, some of the highlighted keywords appear in those topics. For instance, in Topic $3$ and Topic $6$, physical evidences such as bullet, DNA, and saliva that are likely found at the scene are identified and clustered, 
which can help CIP officials identify the appearance of physical evidence. Semantic NMF with keyword highlighting results can show CIP officials what details they can expect from the collection of documents such as different crime types or evidence, which can potentially accelerate the decision-making. Comparing results from the two semantic NMF methods, keyword highlighting leads to more interpretable and demonstrably better topics for the AOB data set.

\begin{table}[htb]
    \centering
    \resizebox{\textwidth}{!}{
    \begin{tabular}{cccccccccc}
    \Xhline{2\arrayrulewidth}
		\textbf{Topic 1} & \textbf{Topic 2} & \textbf{Topic 3} & \textbf{Topic 4} & \textbf{Topic 5} & \textbf{Topic 6} & \textbf{Topic 7}& \textbf{Topic 8}& \textbf{Topic 9} & \textbf{Topic 10}\\
	\hline 
appellant & stopped & involving & going & primary & help & instruct & improper & member & transcript \\
people & identified & object & imprisonment & police & drove & plaintiff & perpetrator & anything & without \\
evidence & violent & minutes & imposed & photographs & verdicts & including & self & established & regarding \\
trial & past & mind & daughter & photo & however & imposed & including & instant & inadmissible \\
jury & introduced & propensity & passenger & never & including & might & actions & separate & proof \\
would & turned & plus & brief & next & asked & specifically & holding & establish & word \\
supra & arrived & generally & plaintiff & caliber & error & notice & accused & failure & respondent \\
murder & estrada & identity & acts & position & noted & asked & thus & verdict & present \\
also & gang & whether & timely & committing & certain & dated & honorable & discretion & plus \\
years & legal & come & answer & arguments & ibid & related & included & provides & victim \\
    \Xhline{2\arrayrulewidth}
    \end{tabular}
    }
    \caption{The top 10 keywords learned by semantic NMF on the AOBs from both categories}
    \label{tab:semantic_aob}
\end{table}

\begin{table}[htb]
    \centering
    \resizebox{\textwidth}{!}{
    \begin{tabular}{cccccccccc}
    \Xhline{2\arrayrulewidth}
		\textbf{Topic 1} & \textbf{Topic 2} & \textbf{Topic 3} & \textbf{Topic 4} & \textbf{Topic 5} & \textbf{Topic 6} & \textbf{Topic 7}& \textbf{Topic 8}& \textbf{Topic 9} & \textbf{Topic 10}\\
	\hline 
appellant & snitch & caliber & semen & executed & saliva & ligature & notice & mixture & submitted \\
people & informant & fired & period & follows & sperm & around & timely & dna & certificate \\
evidence & lineup & bullet & caused & fully & dna & agreed & filed & least & authorities \\
trial & provides & recovered & might & addressed & inside & allegation & january & consistent & word \\
jury & shown & shots & exhibit & clerk & object & shot & amended & excluded & appealability \\
supra & photo & body & envelope & perjury & get & presented & february & blood & dated \\
would & simply & scene & injury & warrant & analysis & help & december & juror & document \\
murder & eyewitness & direction & document & envelope & reference & killed & judgment & recovered & respectfully \\
apartment & imprisonment & weapon & addressed & foregoing & taken & go & motion & ballistics & words \\
penal & photographs & feet & fully & february & apartment & talk & abstract & taken & statement \\
    \Xhline{2\arrayrulewidth}
    \end{tabular}
    }
        \caption{The top 10 keywords learned by semantic NMF with keyword highlighting on the AOBs from both categories}
    \label{tab:semantic_key_aob}
\end{table}

\FloatBarrier
\subsubsection{HNMF} \label{subsubsec: HNMF_AOB}
For both the top-down (Section \ref{subsubsec:TD_HNMF_method}) and bottom-up (Section \ref{subsubsec:BU_HNMF_method}) HNMFs, we begin from an initial factorization of rank 10. For top-down HNMF, each of the 10 super-topics in the first layer is split into three sub-topics. As a result, we discovered a total of 30 sub-topics from AOBs. While, in the bottom-up HNMF, the 10-topic layer is regarded as the sub-topic layer, where 2 super-topic layers are built to combine the 10 sub-topics subsequently. 

In Table \ref{tab:HNMF_topdown_AOB}, we display the topic keywords in each of the 2 layers formed by top-down NMF on the AOBs. These topics and the hierarchical structure are interesting and meaningful to explore. Super-topic 3 concerning murder, for example, is split into 3 more specific sub-topics: the first concerning premeditated murder, the second concerning aiding and abetting, and the third concerning gang related murder. Super-topic 4 related to ``burglary" is branched into 3 sub-topics, one involving specific charges, one describing case details, and the other specifying injuries. 

In Tables \ref{tab:hnmf_aob_10} through \ref{tab:hnmf_aob_2} we display a three layer bottom-up HNMF on the AOBs with topic numbers 10, 4, and 2. The hierarchical factorization at the second layer combines the layer one topics related to murder (topics 1 through 3), burglary (topics 4 and 5), and some of the topics related to gangs and courtroom trials (topics 8 through 10). The third layer of hierarchical factorization then combines together topics at the second layer related to murder and burglary (topics 1 and 2).

\begin{table}[htb]
    \centering
    \resizebox{\textwidth}{!}{
    \begin{tabular}{p{5cm}|p{4cm}p{4cm}p{4cm}}
    \Xhline{2\arrayrulewidth}
		\textbf{Primary Topics} & \textbf{Subtopic 1} & \textbf{Subtopic 2} & \textbf{Subtopic 3}
\\
\hline 
\textbf{Super-topic 1}: \\ 
prosecutor, misconduct, prejudicial, witness, prejudice, admission, statements, uncharged, prosecutorial, probative
& 
\textit{prosecutor}, witness, \textit{misconduct}, \textit{prejudicial}, \textit{statements}, prejudice, \textit{admission}, discretion, burglary, robbery
& 
gang, prosecutor, members, member, beer, expert, shooting, witness, attempted, officer
& 
instruction, identity, eating, stabbed, tank, wearing, fight, scene, perpetrator, citation
\\
\hline
\textbf{Super-topic 2}: \\ 
gang, members, member, expert, shooting, crips, gangs, enhancement, car, territory
& 
\textit{gang}, \textit{member}, \textit{members}, \textit{shooting}, \textit{expert}, murder, \textit{crips}, enhancement, \textit{gangs}, prosecutor
& 
sex, male, murder, \textit{car}, dna, injury, \textit{shooting}, trigger, robbery, september
& 
\textit{gang}, beer, trunk, \textit{expert}, men, intent, \textit{car}, attempted, \textit{members}, attempt

\\
\hline
\textbf{Super-topic 3}: \\ 
murder, intent, aider, abettor, shooting, premeditation, degree, killing, attempted, premeditated
&
\textit{murder}, \textit{premeditation}, deliberation, \textit{killing}, \textit{shooting}, \textit{intent}, \textit{attempted}, \textit{premeditated}, shot, finding
&
\textit{murder}, \textit{aider}, \textit{abettor}, abetting, aiding, perpetrator, probable, instruction, car, \textit{intent}
&
gang, expert, members, beer, \textit{intent}, member, \textit{shooting}, \textit{murder}, prosecutor, tattoo
\\
\hline
\textbf{Super-topic 4}: \\ 
strike, enhancement, felony, burglary, injury, discretion, strikes, bodily, serious, robbery
&
\textit{strike}, \textit{burglary}, \textit{strikes}, \textit{felony}, convictions, \textit{discretion}, threat, \textit{robbery}, priors, \textit{serious}
&
mayhem, instruction, unanimity, eating, beer, stabbed, identity, citation, \textit{injury}, tank
&
\textit{injury}, \textit{bodily}, \textit{enhancement}, \textit{strike}, enhancements, year, \textit{felony}, \textit{serious}, assault, personally
\\
\hline
\textbf{Super-topic 5}: \\ 
sexual, sex, rape, duress, vagina, touching, lewd, penis, offenses, penetration
& 
\textit{sex}, \textit{sexual}, \textit{rape}, propensity, \textit{offenses}, falsetta, molestation, \textit{penis}, raped, admission
&
\textit{duress}, \textit{lewd}, \textit{sexual}, \textit{penetration}, \textit{touching}, acts, \textit{vagina}, touched, penis, occurred
&
lines, \textit{sexual}, hearsay, aunt, abuse, spontaneous, expert, exam, testify, declarant
\\
\hline
\textbf{Super-topic 6}: \\ 
accomplice, robbery, corroboration, special, burglary, murder, statements, circumstance, instruction, conspiracy
&
\textit{accomplice}, \textit{murder}, \textit{corroboration}, \textit{special}, \textit{robbery}, \textit{circumstance}, apartment, \textit{statements}, instruct, accomplices
&
gang, members, expert, threat, beer, member, attempted, shooting, car, witness
&
\textit{instruction}, abuse, sexual, \textit{robbery}, testify, strike, witnesses, sex, injury, discretion
\\
\hline
\textbf{Super-topic 7}: \\ 
juror, jurors, prospective, misconduct, deliberations, motion, verdict, dna, excused, shooting
&
sexual, abuse, injury, instruction, discretion, strike, bodily, robbery, probation, prosecutor
&
gang, members, member, vargas, \textit{shooting}, murder, expert, car, beer, threat
&
\textit{juror}, \textit{jurors}, \textit{misconduct}, \textit{prospective}, \textit{shooting}, prosecutor, eyewitness, \textit{motion}, \textit{deliberations}, photo
\\
\hline
\textbf{Super-topic 8}: \\ 
car, phone, detective, apartment, officer, shooting, shot, motion, going, plea
&
\textit{car}, \textit{phone}, murder, \textit{detective}, \textit{apartment}, \textit{shooting}, shot, \textit{officer}, robbery, \textit{going}
&
vargas, suv, gang, \textit{car}, blood, men, declaration, pants, bills, partner
&
probation, \textit{plea}, report, request, unknown, per, conditions, violation, november, file
\\
\hline
\textbf{Super-topic 9}: \\ 
suggestive, witness, eyewitness, photo, lineup, shooting, identifications, photographic, suspect, pack
&
\textit{suggestive}, \textit{witness}, \textit{photo}, \textit{eyewitness}, \textit{pack}, \textit{lineup}, \textit{identifications}, \textit{photographic}, \textit{suspect}, procedure
&
instruction, eating, stabbed, identity, tank, wearing, fight, scene, assailant, perpetrator
&
gang, expert, lineup, \textit{suggestive}, beer, members, \textit{shooting}, member, car, special
\\
\hline
\textbf{Super-topic 10}: \\ 
passion, manslaughter, heat, voluntary, instruction, provocation, self, instruct, lesser, murder
&
\textit{passion}, \textit{manslaughter}, \textit{heat}, \textit{voluntary}, \textit{murder}, \textit{provocation}, \textit{lesser}, \textit{instruct}, \textit{self}, malice
&
\textit{instruction}, injury, phone, prosecutor, robbery, strike, testify, bodily, weapon, dna
&
gang, threat, beer, expert, members, member, threats, attempted, car, associate
\\
\hline
\end{tabular}
}
\caption{The top $10$ keywords learned by Top-down HNMF on the AOBs from both categories. Keywords in primary topics are italicized in subtopics.}
\label{tab:HNMF_topdown_AOB}
\end{table}

\FloatBarrier

\begin{table}[htb]
    \centering
    
    \resizebox{\textwidth}{!}{
    \begin{tabular}{cccccccccc}
    \Xhline{2\arrayrulewidth}
		\textbf{Topic 1} & \textbf{Topic 2} & \textbf{Topic 3} & \textbf{Topic 4} & \textbf{Topic 5} & \textbf{Topic 6} & \textbf{Topic 7}& \textbf{Topic 8}& \textbf{Topic 9} & \textbf{Topic 10}\\
	\hline 
murder & murder & car & burglary & strike & sexual & prosecutor & suggestive & juror & gang \\
manslaughter & accomplice & phone & apartment & enhancement & sex & misconduct & witness & jurors & members \\
passion & aider & detective & intent & injury & rape & witness & eyewitness & prospective & member \\
voluntary & abettor & shooting & mayhem & felony & duress & prejudicial & photo & misconduct & expert \\
heat & robbery & apartment & residential & bodily & vagina & statements & lineup & deliberations & shooting \\
killing & special & murder & threat & robbery & touching & prejudice & identifications & motion & crips \\
premeditation & intent & shot & instruction & discretion & lewd & admission & shooting & verdict & gangs \\
malice & abetting & officer & felony & strikes & penis & prosecutorial & photographic & dna & murder \\
provocation & aiding & going & burglaries & imposed & offenses & objection & suspect & excused & enhancement \\
deliberation & instruction & got & unanimity & firearm & penetration & errors & pack & instruction & car \\
    \Xhline{2\arrayrulewidth}
    \end{tabular}
    }
    \caption{The top $10$ topic keywords learned by the first layer of bottom-up HNMF on AOBs from both categories}
    \label{tab:hnmf_aob_10}
\end{table}

\begin{figure}[h]
\begin{minipage}{0.55\textwidth}
    \centering
   
    \begin{tabular}{cccc}
    \Xhline{2\arrayrulewidth}
		\textbf{Topic 1} & \textbf{Topic 2} & \textbf{Topic 3} & \textbf{Topic 4}\\
	\hline 
murder & burglary & sexual & gang \\
car & strike & sex & members \\
shooting & felony & rape & member \\
phone & robbery & duress & shooting \\
shot & enhancement & prosecutor & expert \\
detective & apartment & vagina & prosecutor \\
instruction & intent & touching & juror \\
degree & convictions & lewd & murder \\
apartment & discretion & offenses & witness \\
killing & serious & penis & crips \\
    \Xhline{2\arrayrulewidth}
    \end{tabular}
     \captionof{table}{The top $10$ topic keywords learned by the second layer of bottom-up HNMF on AOBs from both categories} 
    \label{tab:hnmf_aob_4}
\end{minipage}
\hfill
\begin{minipage}{0.4\textwidth}
    \centering
    
    \begin{tabular}{cc}
    \Xhline{2\arrayrulewidth}
		\textbf{Topic 1} & \textbf{Topic 2}\\
	\hline 
murder & gang \\
car & members \\
shooting & member \\
instruction & shooting \\
prosecutor & expert \\
burglary & prosecutor \\
apartment & juror \\
phone & murder \\
robbery & witness \\
detective & crips \\
    \Xhline{2\arrayrulewidth}
    \end{tabular}
    \captionof{table}{The top $10$ topic keywords learned by the third layer of bottom-up HNMF on AOBs from both categories} 
    \label{tab:hnmf_aob_2}
\end{minipage}
\end{figure}

\FloatBarrier
\subsection{Classification} \label{subsec:classification}

Besides investigating the hidden topics from the initial letters and the AOBs, we take advantage of the supervised nature of SNMF (Section \ref{subsubsec: SNMF_method}) and SSNMF (Section \ref{subsubsec: SSNMF_method}) to provide some useful insights into classification questions raised by CIP officials. We train models to reconstruct decision type labels from initial letter data and crime type labels from AOB data. Labeling decision type can help to determine whether a new case is worth pursuing by examining only the initial letters. In practice, if the predicted decision type for a case is ``\textit{Cases for Investigation}", then this case may be worth pursuing because it is classified as similar to cases that are ready for additional investigation. Labeling crime type can provide an overall understanding of each case before officials take a closer investigation. For each of the following experiments, we calculate the average LAS score (\ref{subsubsec:LAS}) over $10$ trials with different training and testing data partitioned by $75\%$ to $25\%$ ratios to measure algorithms' classification accuracy. For building the tf-idf vocabulary, we use the tuning parameters ``max\_df=$0.8$" and ``min\_df=$0.2$" in the function ``TfidfVectorizer".  

\FloatBarrier
\subsubsection{Decision Type Classification}
In addition to constructing the training and testing data matrices, we manually create the corresponding decision type label matrices described as follows. Since one case can only belong to one decision type, for each case in the training or testing data, the corresponding column of the label matrix is either $[1, 0]^T$ to indicate that this case belongs to the \textit{Cases for Investigation} type or $[0, 1]^T$ to indicate it belongs to \textit{Cases to be closed}. The SNMF algorithm yields an average LAC score of $65\%$ while the the SSNMF algorithm yields an average LAC score of $55\%$. The higher rate in SNMF is counter-intuitive since SNMF incorporates fewer cases than SSNMF when training the model. However, the low LAC score of SSNMF can probably be explained by the fact that our training data set is small and adding more cases will lead to over-fitting issues.  

\FloatBarrier
\subsubsection{Crime Type Classification}

\begin{figure}[h!]
  \begin{subfigure}{16cm}
    \centering
        \includegraphics[width=16cm]{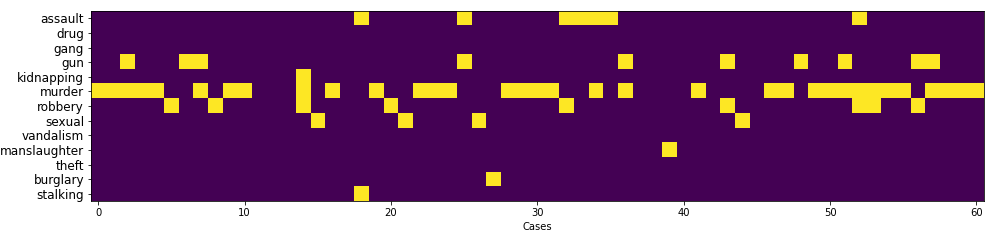}
        \caption{True crime labels for cases in testing data from both categories.}
        \label{fig:class_crime_test}
  \end{subfigure}
  
  \begin{subfigure}{16cm}
    \centering
        \includegraphics[width=16cm]{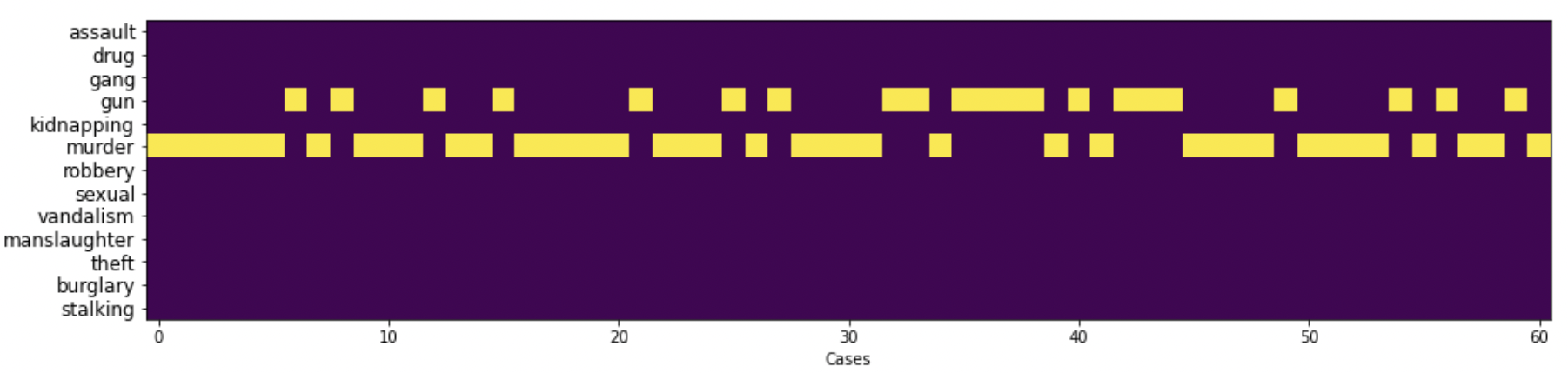}
        \caption{Reconstructed crime labels for cases in testing data by SNMF from both categories.}
        \label{fig:class_crime_test_recon}
  \end{subfigure}
 
  \begin{subfigure}{16cm}
    \centering
        \includegraphics[width=16cm]{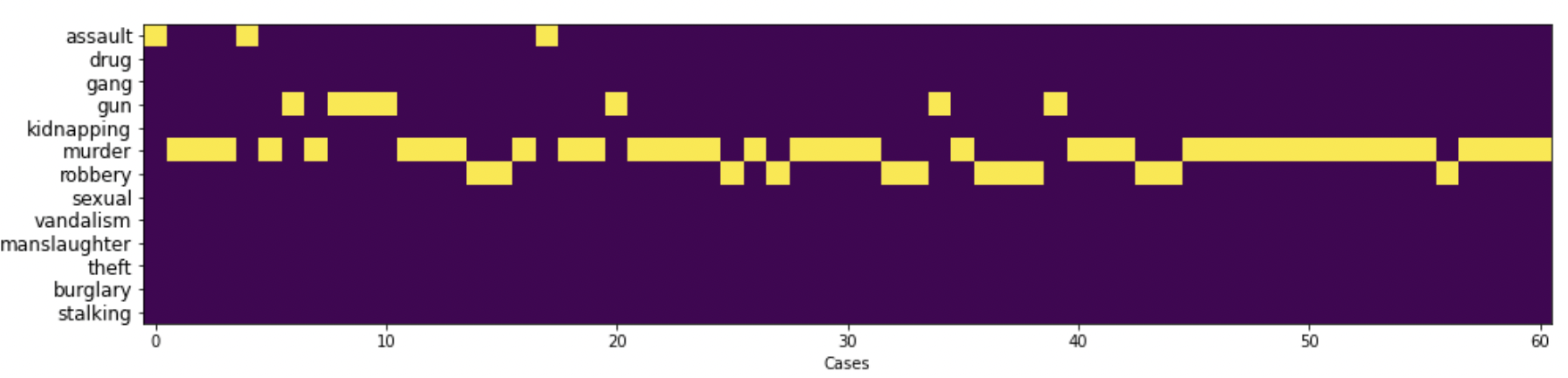}
        \caption{Reconstructed crime labels for cases in testing data by SSNMF from both categories.}
        \label{fig:class_race_test_recon_ssnmf}
  \end{subfigure}
   \caption{Comparison of original labels and reconstructed ones. The yellow pixel indicates that the case is assigned to the corresponding crime label on the y-axis, while the dark purple pixel indicates that the case is not assigned.}
\end{figure}

We extract crime labels of each case as described in Section \ref{subsec:labels} and then implement both algorithms. SNMF algorithm results in an average LAC score of $92\%$,
while the SSNMF algorithm yields an average LAC score of $91.8\%$. Both resulting label matrices are heavily centered towards the dominant labels with the less frequent labels having coefficients very close to 0. Overall, the two algorithms generate promising prediction rates for crime labels, but it can be possibly due to the data set being highly biased towards the crime of murder. Further study on a larger data set is needed to more accurately determine the value of crime type classification. To visualize the results of crime type classification, Figure \ref{fig:class_crime_test} shows the actual crime labels in the testing data. Figure \ref{fig:class_crime_test_recon} shows the reconstructed crime labels by SNMF, and Figure \ref{fig:class_race_test_recon_ssnmf} shows the reconstructed crime labels by SSNMF.

\FloatBarrier
\section{Discussion and Future Works} \label{sec:Conclusion}
In this paper, we first provide an exposition of popular variants of Non-negative Matrix Factorization. Then, we discover and analyze meaningful lexical topics from the initial letters and AOBs provided by the California Innocence Project (CIP) through the various NMF methods. We also reconstruct crime type and decision type labels for each case file using (semi)supervised Matrix Factorization methods.

In general, topics generated from the initial letters (discussed in Section \ref{subsec:Result_IL}) cover three major themes: seeking assistance, trial information, and evidence. Across all methods, we consistently observe a topic related to seeking assistance, which is the major purpose of the initial letters.
Topic results generated from classical NMF are generally trial-related and thus can help CIP identify cases whose initial letters describe their trials comprehensively at first hand. Semantic NMF with keyword highlighting reveals topics related to evidence such as eyewitness, fingerprints, and video footage, which allows CIP to know about the appearance of detailed physical evidence in the document before reading them. Results from top-down and bottom-up HNMF reveal the hidden hierarchical structure, which can potentially help CIP specify or classify trial and evidence information. Identifying the trial information and the corresponding evidence, our analysis of initial letters offers CIP a brief overview of cases. 

Topics generated from the AOBs (discussed in Section \ref{subsec: Topics_AOB}) cover two major themes: type of crime and physical evidence. Topic results of classical NMF reveal the diversity of crime types in the AOB documents. Analyzing cases within those topics can potentially help CIP identify special cases belonging to multiple crime types. Top-down HNMF provides additional information and classification about one type of crime, while the bottom-up approach shows how certain crimes can be combined to form a more general category. Besides crime types, CIP also focuses on crucial facts and potential evidence stated in the AOB documents. Semantic NMF with keyword highlighting, by adding more weight on keywords related to physical evidence, generates topics covering various types of evidence. Combining both pieces of knowledge about crime types and physical evidence, those topics can help CIP gain basic understanding (before reading) of the information included in each case, and this pre-knowledge can increase their efficiency in evaluating the case. While results from AOBs are very informative, results from initial letters are also valuable as they depict a general picture of cases from various perspectives such as trial information and evidence when CIP first screens them.

For both initial letters and AOBs, semantic NMF itself generates overlapping and hard-to-interpret topics. A potential explanation is that the extra contextual information semantic NMF adds to the topics does not provide additional information for understanding. Since the SPPMI matrix is trained from the CIP data set itself, bias is unavoidable. In order to reduce the bias, we could train our SPPMI matrix from different data corpus and compare the results. However, its idea of capturing semantic information motivates us to propose a novel keyword highlighting version of semantic NMF. Through emphasizing important keywords (see \ref{subsec:keywords Highlighted}), this method encourages more topics related to evidence in both initial letters and AOBs. As CIP officials input different sets of highlighting keywords, the topic modeling result will vary correspondingly. As a result, the flexible nature of semantic NMF with keyword highlighting enables CIP officials to gain more control over what kind of information they can expect from case files. Thus, semantic NMF with keyword highlighting could potentially be the most effective method to assist CIP's decision-making due to its flexibility in both highlighting keywords and the SPPMI matrix.
 
Our second objective is to classify and predict the label of each case using (semi)supervised NMF (discussed in Section \ref{subsec:classification}). We perform two classification tasks: the first one on initial letters deciding whether a case should go into investigation or be closed (decision type), and the second one on AOBs deciding the types of crime concerning the case (crime type). Our results have an average prediction accuracy score of $60\%$ for the decision types and $90\%$ for the crime types. The classification accuracy for decision types may suffer from over-fitting due to the limited size of the data set, so more research must be performed on a larger data set to help us draw the conclusion. Potentially, we wish that the first classification can help simplify the screening process since CIP officials can first examine the cases that are classified to be similar to past cases that were investigated. The model trained for the second classification can help CIP officials  discern the crime type for each case file when there is a large number of incoming case files. 
The decreased classification accuracy in predicting decision types may imply that the process of decision making is too complicated to be handled by machine learning algorithms, such as NMF. Aside from those topic modeling results, human judgement is still crucial and decisive. The number of topics in NMF is decided by the user. When, in reality, there are more topics then we asked for, NMF could potentially disregard some of the minor topics or just represent them with only a few words (eg. the topic with Spanish words disappeared in AOB results). In the case of analyzing important legal documents, ignoring details could affect holistic decision making. Meanwhile, people could be wrong and biased against certain details as well but NMF offers an alternative perspective to the data set. Combining both perspectives from machine learning and human judgment could accelerate the process of making a holistic decision.

Therefore, the purpose of machine learning methods, in our case, is more informative rather than conclusive when it comes to decision-making. While applying those methods to other data sets, users should pay attention to privacy and potential bias. Sensitive variables related to people's personal information shall be removed, such as names, addresses, etc. Bias is a constant topic in machine learning algorithms, such as word embeddings \citep{bias_article}. The algorithms should not take people's gender, race, religion, or different dialects they speak into account; To mitigate those biases while preserving the properties of the word embedding, we should also consider some debiasing methods \citep{bias_paper, bias_article}.
See also \citep{blodgett-etal-2020-language} for a nice discussion of the benefits and pitfalls of such debiasing approaches.

Across all the topic modeling results generated from initial letters, we constantly observed topics consisting of Spanish words. Although the way we pre-process the initial letter data set and the implementation of NMF variants is not able to handle multilingual data, these topics of Spanish stop word topics could help CIP notice the existence of multiple languages in the initial letters. In the future, to better include those initial letters in Spanish, or even other languages, into Topic Modeling algorithms, we could apply Multilingual Unsupervised and Supervised Embeddings (MUSE), which take into account the polysemy of words, to translate them into English \citep{conneau2017word}. Given the syntax complexity of many languages, we hope to capture contextual phrases instead of a single word to increase the interpretability of the resulting topics. We can achieve this by utilizing n-grams instead of mono-grams when building tf-idf vocabularies. Since, some important keywords may be short phrases, applying n-grams can generate keyphrases so that potentially improve our model's flexibility in capturing meaningful topics. We also wish to apply Non-negative Tensor Factorization methods to the CIP data to honor the often multidimensional structure of the data. For example, utilizing time information of each case can help CIP understand the distribution of cases chronologically. As CIP constantly receives new cases over time, we plan to update our topic clustering results utilizing information from those new cases via an online version of NMF.

\section*{Acknowledgments}
The authors appreciate Prof. Elizaveta Rebrova and Dr. Denali Molitor for their guidance in this project. The authors also appreciate the support from UCLA Computational and Applied Math REU, NSF BIGDATA $\#1740325$ and NSF DMS $\#2011140$.



\bibliography{collectedbib.bib}

\appendix
\section{Appendix}

\subsection{Highlighted Keywords } \label{subsec:keywords Highlighted}
eyewitness, microscopy, shaken baby syndronme, sbs, abusive head trauma, aht, false confession, coerced confession, rampart, comparative bullet lead analysis, cbla, tool mark, toolmark,tread mark comparison, tread mark analysis, fiber comparison, impression comparison, impression analysis, arson, ballistics, blood splatter, handwriting comparison, informat, snitch, strangle, ligature, sodomy, sexual assault, intercourse, digital penetration, penetration by a foreign object, saliva, semen, sperm, amylase, seminal fluid, duct tape, bindings, mixture, bite mark, bitemark, fingerprint, fingernail scrapings, shell casing, blood type

\end{document}

%% file: math_commands.tex
\usepackage{amsmath,amsfonts,bm}

\def\eqref#1{equation~\ref{#1}}

\def\1{\bm{1}}

\DeclareMathAlphabet{\mathsfit}{\encodingdefault}{\sfdefault}{m}{sl}
\SetMathAlphabet{\mathsfit}{bold}{\encodingdefault}{\sfdefault}{bx}{n}

\newcommand{\R}{\mathbb{R}}



%% file: header.tex
\usepackage{subcaption}
\usepackage{mathtools}
\usepackage{microtype}
\usepackage{bbm}
\usepackage{makecell}
\usepackage[capitalise]{cleveref}

\usepackage{amsfonts}
\usepackage{amsmath}
\usepackage{IEEEtrantools}
\usepackage{amssymb} 

\usepackage{algorithm}
\usepackage[noend]{algpseudocode}

\usepackage{xcolor}
\usepackage{color}
\usepackage{graphicx}

\newcommand{\norm}[1]{\left\lVert#1\right\rVert}      